\documentclass{article}
\usepackage{spconf,amsmath,graphicx}
\usepackage[dvipsnames]{xcolor}
\usepackage{cite}
\usepackage{url}

\title{Deform-GAN:An Unsupervised Learning Model for Deformable Registration}
\name{Xiaoyue Zhang*, Weijian Jian, Yu Chen, Shihting Yang}
\address{}
\begin{document}
%
\maketitle
\begin{abstract}
Deformable registration is one of the most challenging task in the field of medical image analysis, especially for the alignment between different sequences and modalities. In this paper, a non-rigid registration method is proposed for 3D medical images leveraging unsupervised learning. To the best of our knowledge, this is the first attempt to introduce gradient loss into deep-learning-based registration. The proposed gradient loss is robust across sequences and modals for large deformation. Besides, adversarial learning approach is used to transfer multi-modal similarity to mono-modal similarity and improve the precision. Neither ground-truth nor manual labeling is required during training. We evaluated our network on a 3D brain registration task comprehensively. The experiments demonstrate that the proposed method can cope with the data which has non-functional intensity relations, noise and blur. Our approach outperforms other methods especially in accuracy and speed.
\end{abstract}
\section{Introduction}
\label{sec:intro}
Due to the complex relationship between intensity distributions, multi-modal registration \cite{heinrich2012mind} remains a challenging topic. Optimization based registration, which optimizes similarity across phases or modals by aligning the voxel pairs, has been a dominant solution for a long time. However, along with the high complexity of solving 3D images optimization, it is very hard to define a descriptor which is robust enough to cope with the most considerable differences between the image pairs.

Nowadays, a lot of methods leveraging deep learning has been proposed to solve the problems mentioned above. These approaches usually require registration fields of ground truth or landmarks which need to be annotated by experts. Some methods \cite{balakrishnan2018unsupervised}\cite{dalca2018unsupervised} explored unsupervised strategies built on the spatial transformer network.

There are two main challenges in unsupervised-learning based registration. The first one is to define a loss which can efficiently provide similarity measurement across modalities or sequences. For example, mutual information (MI) has been widely and successfully used in registration tasks. But it requires binning or quantizing, which can cause gradient vanishing problem\cite{lau2019unsupervised}. The second challenge is no ground-truth. The intuition to solve multi-modal problem is image-to-image translation\cite{hu2018adversarial}\cite{fan2018adversarial}. But without pixel-wise aligned data pairs, it is difficult to train a GAN to generate synthesized images in which the all texture is mapping to the source exactly. For example, Cycle-GAN can generate the images from MR which just look like CT, but the accuracy in the details cannot meet the requirements of registration.

In this paper, we propose a novel unsupervised method which can easily achieve deformable registration between different sequences or modalities. Local gradient loss, an efficient and robust metric, is the first time to be used in deep-learning-based registration method. We combine adversarial learning approach with spatial transformation to simplify multi-modal similarity to mono-modal similarity. Experiment results show that our approach is competitive to state-of-the-art image registration solutions in terms of accuracy and speed.

\section{Method}
\label{sec:format}
\begin{figure}[htb]
\begin{minipage}{1.0\linewidth}
  \centering
  \centerline{\includegraphics[width=8.5cm]{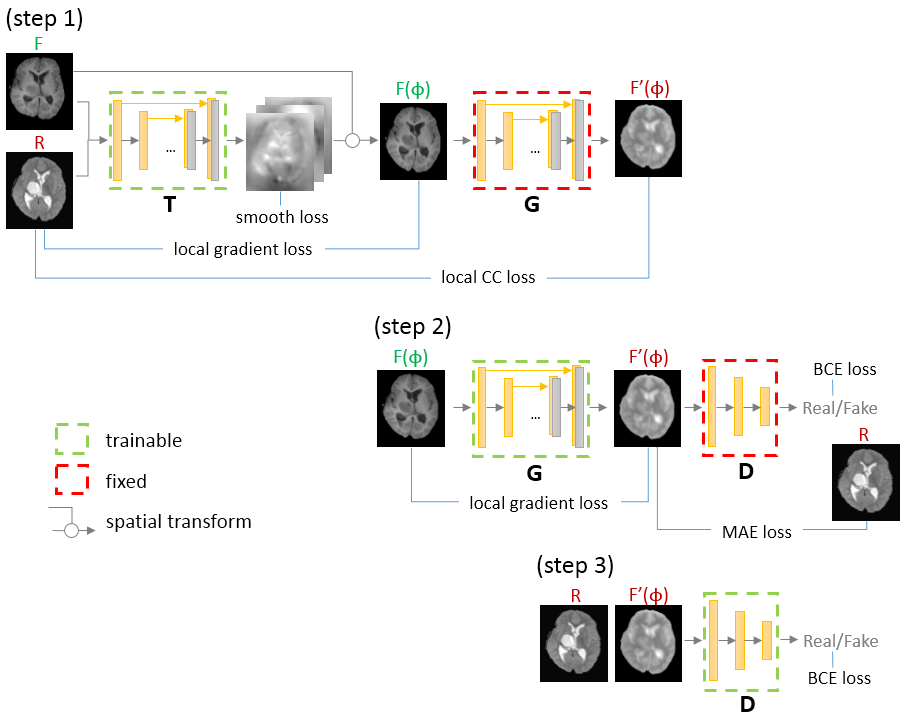}}
\end{minipage}
\caption{Overview architecture of the proposed model and training steps. Our model mainly consists of three components: a transformation network $T$, a generator $G$ and a discriminator $D$. While training, we take one gradient descent step on $T$, one step on $G$ and one step on $D$ by turns.}
\label{}
\end{figure}
Our model mainly consists of three parts: an image transformation network $T$ which outputs the registration warp field for spatial transformation, an image generator $G$ which does multi-modal translation and a discriminator $D$ which can distinguish real images and synthesized images. The architecture of our model and the training details is illustrated in Fig.1.

\subsection{Architecture}
\label{ssec:subhead}
The transformation network $T$ takes the reference image $R$ and the floating image $F$ as input, and then outputs the registration warp field $\phi$ . The mapping can be written as $T:(R,F)\Rightarrow\phi$. The floating image $F$ is warped into $F(\phi)$ using a spatial transformation function. Then $F(\phi)$ is sent to the generator $G$ which synthesizes images $F’(\phi)$ of the reference image domain. That, in turn, provides an easier registration task of single-modal between $R$ and $F’(\phi)$.

In the proposed network, registration problem is divided into two parts: multi-modal registration $(R,F(\phi))$ and mono-modal registration $(R,F’(\phi))$, which share the same deformation warp field $\phi$. So every voxel $F’(\phi(p))$ in synthesized images should be mapped to $F(\phi(p))$ precisely. However, in the early learning period, $T$ is poor and the registration result is not accurate. If we use the architecture like Pix2Pix\cite{isola2017image} and send the unpaired $F(\phi)$ and $R$ to the discriminator, the generator will be confused and generate a misaligned $F’(\phi)$. To solve this problem, we present a gradient-constrained GAN method for unpaired data. This method is different in that the loss is learned, and can, not only penalize any possible structure that differs between output and target, but also penalize that between output and source. The task of generator consists of three parts: fooling the discriminator, minimizing $L1$ distance between output and the target, and keeping the output texture similar to the source. The discriminator’s job remains unchanged: only to discriminate real and fake.

Both T and G are U-Net-like\cite{ronnebergerconvolutional} network. For details, our code and model parameters are available online at \url{https://github.com/Lauraxy/Multi_Modal_Registration}. These three networks should be trained by turns: one step of optimizing $D$, one step of optimizing $G$ and one step of optimizing $T$. Note that when training one network, the weights of other two networks should be fixed. Please refer to Fig.1 for details. As $G$ is updated gradually, $F’(\phi)$ becomes more and more real which helps to update $T$ and $\phi$. Then $F(\phi)$ can be aligned better to $R$ and in turn contributes to train $G$. This results in that $T$ and $G$ are reaching mutual beneficial.

\subsection{Loss}
\label{ssec:subhead}
We had tried several loss functions for evaluating similarity between multi-sequence images, such as MI, NGF, and so on. However, each of them has their own weakness and cannot achieve satisfying registration results. For example, we try using Parzen Window estimation of MI to solve gradient-vanish problem, but huge memory consumption make it difficult to train a model in practice. NGF, in our experiment, cannot drive the warp field to convergence. Here we present a local gradient loss which can depict local structure information across modalities. It is similar to NGF, but more robust against noise and easy to converge fast.

Suppose that $p$ is a voxel position of volume $I$, and we can get the local gradient by:
\begin{equation}
\nabla\hat{I}(p)= (\sum_{p\in{n^3}} x'(p), \sum_{p\in{n^3}} y'(p), \sum_{p\in{n^3}} z'(p))
\end{equation}

Where $x’$, $y’$, $z’$ are gradient filed and $p$ iterates over a $n^3$ volume around $p$. Then the gradient can be normalized by:
\begin{equation}
n(I,p)=\frac{\nabla\hat{I}(p)}{\lVert\nabla\hat{I}(p)\rVert+\varepsilon}
\end{equation}

Where $\lVert\cdot\rVert$ means L2 distance. The local gradient loss between $R$ and $F$ can be defined as follow:
\begin{equation}
L_{LG}(R,F)=\sum_{p\in{\Omega}}|n(R,p) \cdot n(F,p)|
\end{equation}

$\Omega$ is the volume domain of $R$ and $F$. In the experiment of local gradient, if $n$ in Eq.1 is too small, the network would be difficult to converge. Instead, if $n$ is too large, the edge of $R$ and $F$ cannot be aligned accurately. Finally we set $n=7$ and get the best results. 

Next we will talk about the loss of T, which can be expressed as:
\begin{equation}
{L}_{T}(R,F,\phi)={L}_{sim}(R,F(\phi))+\alpha{L}_{smooth}(\phi)
\end{equation}

We set $L_{sim}$ as two parts: the negative local cross-correlation of $R$ and $F'(\phi)$,  the negative local gradient distance between $R$ and $F(\phi)$:
\begin{equation}
{L}_{sim}(R,F(\phi))=-L_{LCC}(R,F'(\phi))-\beta L_{LG}(R,F(\phi))
\end{equation}

Smooth loss, which enforce spatially smooth deformation, can be set as follow [2]:
\begin{equation}
{L}_{smooth}(\phi)=\sum_{p\in{\Omega}}{\lVert \nabla\phi(p) \rVert}^2
\end{equation}

Then we talk about the generator $G$ and discriminator $D$. First of all let us review Pix2Pix, a promising approach for many image-to-image translation tasks. The loss of Pix2Pix can be expressed as:
\begin{equation}
{L}_{G^*}=arg \min \limits_{G} \max \limits_{D} {L}_{c^{GAN}} (G,D)+\lambda L_{L1}(G)
\end{equation}

Where ${L}_{c^{GAN}}$ is the objective of a conditional GAN\cite{isola2017image}, and $L_{L1}$ is the L1 distance between the source and the ground truth target. Different from Pix2Pix, in multi-modal registration task, the source and the target are not pixel-wise mapping data. That means directly push the source to near the ground truth in an L1 sense may lead to false translation, which is harmful for registraion. Here we introduce the local gradient loss to constrain gradient distance between the synthesized images $F'(\phi)$ and the source images $F(\phi)$ and keep the output texture similar to the source. We mix the GAN objective with local gradient loss to a complete loss:
\begin{equation}
\begin{aligned}
L_{G'}=& arg \min \limits_{G} \max \limits_{D} {L}_{c^{GAN}} (G,D) - \mu L_{LG}(F'(\phi),F(\phi))\\ & + \lambda {L}_{L1} (F'(\phi),R)
\end{aligned}
\end{equation}

\section{Experiments and Results}
\label{sec:pagestyle}

\subsection{Dataset}
\label{ssec:subhead}
We evaluated our method with Brain Tumor Segmentation (BraTS) 2018 dataset[12], which provides a large number of multi-sequence MRI scans, including T1, T2, FLAIR, and T1Gd. Different sequence in the dataset have been aligned very well. We evaluated the registration on T1 and T2 data pairs. We randomly chose 235 data for training and the rest 50 for testing. We cropped and downsized the images to the input size of $112\times128\times96$. We added random shift, rotation, scaling and elastic deformation to the scans and generated data pairs for registration, while the synthetic deformation fields can be regarded as registration ground-truth. The range of deformations can be large enough to -40~+40 voxels and it is a challenge of registration. 

\subsection{Baseline Methods}
\label{ssec:subhead}
We compare two well-established image registration methods with ours: A conventional MI-based approach[?] and VoxelMorph method[4]. In the former one, MI is implemented as driving forces within a fluid registration framework. The latter one introduces novel diffeomorphic integration layers combined with a transform layer to enable unsupervised end-to-end learning. But the original VoxelMorph set similarity as local cross-correlation which only function well in single-model registration.As described in chapter 2.2, we also tried several similarity metric with Voxel-morph framework, such as MI, NGF and LG. But only LG is capable of the regis-tration task. So we just use Voxelmorph with LG for comparison. 

\subsection{Evaluation}
\label{ssec:subhead}
We set the loss weight as: $\alpha=1$in Eq.4, $\beta=2$ in Eq.5 and $\mu=5$, $\lambda=100$ in Eq.8. For CC and local gradient, window size is set as $7\times7\times7$. We use ADAM optimizer with learning rate $1e-4$. NVIDIA Geforce 1080ti GPU with 11GB memory is applied for training and testing. To evaluate the effect of gradient-constrained loss(Eq.8) in generator $G$, we set the network with and without gradient-constrained, named Deform-GAN-2 and Deform-GAN-1, respectively.
\begin{figure}[htb]
\begin{minipage}{1.0\linewidth}
  \centering
  \centerline{\includegraphics[width=8.5cm]{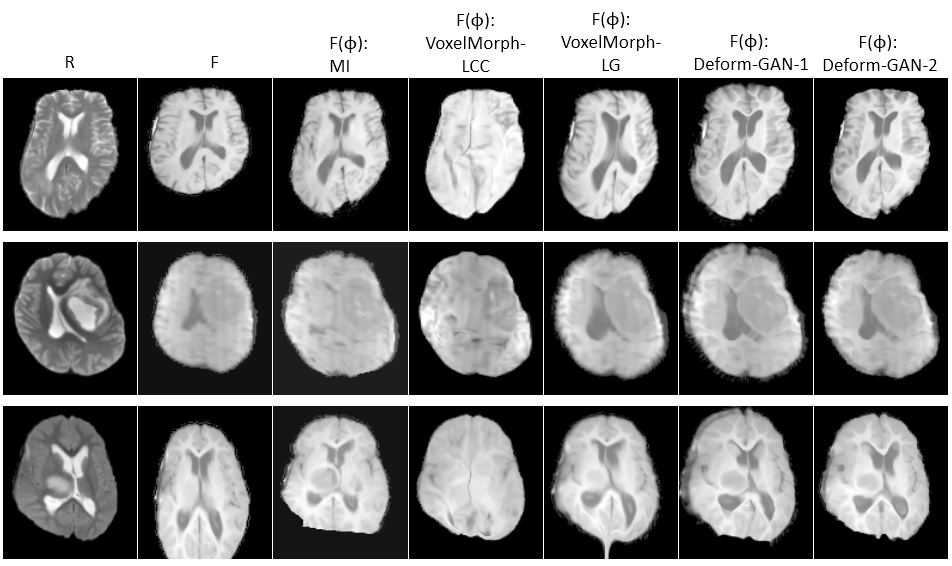}}
\end{minipage}
\caption{Registration results for different methods.}
\label{}
\end{figure}

The registration results are illustrated in Fig.2. MI is intrinsically a global measure and so its local estimation is difficult. VoxelMorph with local CC is based on gray value and cannot function well for cross-sequence. The results of VoxelMorph with gradient loss become much better and can handle large deformation between $R$ and $F$. This demonstrates the effectiveness of the local gradient loss. Our methods, both Deform-GAN-1 and Deform-GAN-2, prove higher accuracy of registration. Even for blur and noisy image (please see the second row), Deform-GAN can get satisfying results, and obviously, Deform-GAN-2 is even better.

For further evaluation on the two setting of Deform-GAN, the warped floating images $F(\phi)$ and synthesized images $F'(\phi)$ from two GANs at different stages of training are shown in Fig.3. We can see that gradient constraint brings the faster convergence during training. Even at first epoch, white matter can be seen clearly in $F'(\phi)$. What’s more, Deform-GAN-2 is more stable in the training process (As the yellow arrows point out, there is less noise in $F'(\phi)$ of Deform-GAN-2 than that of Deform-GAN-1). Note that $F'(\phi)$ is important for calculating $CC(R,F'(\phi))$, it should be aligned to $F(\phi)$ strictly. The red arrows point out that the alignment between $F(\phi)$ and $F'(\phi)$ of Deform-GAN-2 is much better.
\begin{figure}[htb]
\begin{minipage}{1.0\linewidth}
  \centering
  \centerline{\includegraphics[width=8.5cm]{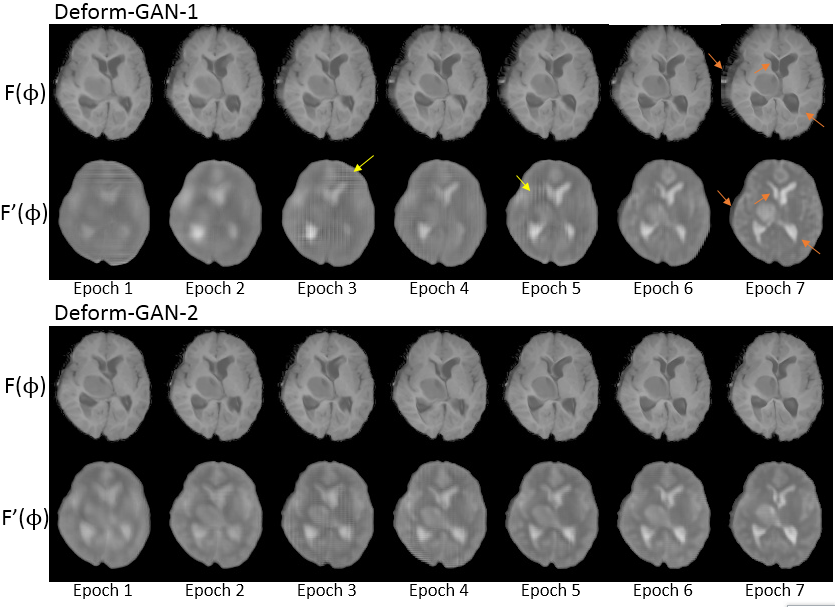}}
\end{minipage}
\caption{Warped floating images $F(\phi)$ and synthesized images $F'(\phi)$ from two Deform-GANs at different stages of training. The yellow arrows point out that Deform-GAN-2 is more stable than Deform-GAN-1. The red arrows indicate the misaligned area between $F(\phi)$ and $F'(\phi)$ in Deform-GAN-1. We can also see that Deform-GAN-2 learns more quickly in the early stages of training.}
\label{}
\end{figure}

In order to quantify the registration results of our methods and the compared methods, we proposed additional evaluation experiments. For the BraTS dataset, we can warp the floating image by synthetic deformation fields as ground truth. Hence, the root-mean-square error (RMSE) of pixel-wise intensity can be calculated for the evaluation. Also, because the mask of tumor is given by BraTS challenge, we can calculate Dice score to evaluate registration result around tumor area. Table 1 shows the quantitative results. It can be seen that our method outperforms the others in terms of tumor Dice and RMSE. In terms of registration speed, deep-learning based methods are significantly faster than the traditional one. In particular, our method only need to run the transformation network  in the inference process, so the runtime is still very fast, though a bit slower than VoxeMorph. 
\begin{table}
\begin{minipage}{1.0\linewidth}
\centering
\caption{Evaluation of registration on the BraTS dataset in terms of RMSE, average Dice of whole tumor and average runtimes on GPU/CPU.}\label{tab1}
\setlength{\tabcolsep}{2.3mm}{
\begin{tabular}{|l|l|l|l|}
\hline
Method & RMSE(\%) &  Tumer Dice & Runtime(s)\\
\hline
MI & 1.39$\pm$0.40 & 0.55$\pm$0.18 & -/6.1   \\

\hline
VoxelMorph-LG & 1.42$\pm$0.36 & 0.61$\pm$0.12 & \textbf{0.09/3.9}  \\

\hline
Deform-GAN-1 & 1.33$\pm$0.31 & 0.67$\pm$0.13 & 0.11/4.4  \\

\hline
Deform-GAN-2 & \textbf{1.18}$\pm$\textbf{0.23} & \textbf{0.69}$\pm$\textbf{0.10} & 0.11/4.4 \\

\hline
\end{tabular}}
\end{minipage}
\end{table}

\section{Conclusion}
\label{sec:typestyle}
A fast multi-modal deformable registration method that makes use of unsupervised learning is proposed. Adversarial learning method combined with spatial transformation helps to reduce similarity calculation between multi-modal to that between mono-modal. We are able to improve the registration results by a weighted sum of local gradient and local $CC$ in a way that the gradient based loss takes global coarse alignment, while local $CC$ loss ensures registration accuracy. Compared to recent learning based methods, our approach can effectively cope with the multi-modal registration problems with large deformation, non-functional intensity relations, noise and blur, promising in state-of-the-art accuracy and fast runtimes.

We declare that we do not have any commercial or associative interest that represents a conflict of interest in connection with the work submitted.

\bibliographystyle{IEEEbib}
\bibliography{refs}
\end{document}